\documentclass[final]{cvpr}
\usepackage{multirow}
\usepackage{times}
\usepackage{epsfig}
\usepackage{graphicx}
\usepackage{booktabs}
\usepackage{caption}
\usepackage{float}
\usepackage{amssymb}
\usepackage{amsmath}
\usepackage[pagebackref=true,breaklinks=true,colorlinks,bookmarks=false]{hyperref}
\newcommand*\samethanks[1][\value{footnote}]{\footnotemark[#1]}

\begin{document}

\title{Interventional Video Grounding with Dual Contrastive Learning}

\author{
Guoshun Nan{$^{1}$$^,$$^*$} \quad Rui Qiao{$^{1}$$^,$}\thanks{Equally contributed}  \quad Yao Xiao{$^{2}$$^,$}\thanks{Work done during internship at StatNLP group, SUTD}  \quad Jun Liu {$^{3}$$^,$}\thanks{Corresponding Authors} \quad Sicong Leng{$^{1}$} \quad Hao Zhang{$^{4}$$^,$$^{5}$} \quad Wei Lu{$^{1}$$^,$}\samethanks \\

\normalsize{$^{1}$} StatNLP Research Group, Singapore University of Technology and Design,
\normalsize {$^{2}$} Shanghai Jiao Tong University, China\\
\normalsize {$^{3}$} Information Systems Technology and Design, Singapore University of Technology and Design, Singapore \\
\normalsize {$^{4}$} School of Computer Science and Engineering, Nanyang Technological University, Singapore\\
\normalsize{$^{5}$ Institute of High Performance Computing, A*STAR, Singapore} \\

{\tt\small \{guoshun\_nan,rui\_qiao\}@sutd.edu.sg, 119033910058@sjtu.edu.cn, jun\_liu@sutd.edu.sg}\\
{\tt\small sicong\_leng@mymail.sutd.edu.sg, zhang\_hao@ihpc.a-star.edu.sg, luwei@sutd.edu.sg}
}

\maketitle

\begin{abstract}
Video grounding aims to localize a moment from an untrimmed video for a given textual query. Existing approaches focus more on the alignment of visual and language stimuli with various likelihood-based matching or regression strategies, i.e., $P(Y|X)$. Consequently, these models may suffer from spurious correlations between the language and video features due to the selection bias of the dataset. 1) To uncover the causality behind the model and data, we first propose a novel paradigm from the perspective of the causal inference, i.e., interventional video grounding (\texttt{IVG}) that leverages backdoor adjustment to deconfound the selection bias based on structured causal model (SCM) and \textit{do}-calculus $P(Y|do(X))$. Then, we present a simple yet effective method to approximate the unobserved confounder as it cannot be directly sampled from the dataset. 2) Meanwhile, we introduce a dual contrastive learning approach (\texttt{DCL}) to better align the text and video by maximizing the mutual information (MI) between query and video clips, and the MI between start/end frames of a target moment and the others within a video to learn more informative visual representations. Experiments on three standard benchmarks show the effectiveness of our approaches. Our code is available on GitHub: \url{https://github.com/nanguoshun/IVG}.
\end{abstract}
\section{Introduction} \label{intro}
\begin{figure}[t]
\begin{center}
\includegraphics[width=0.95\linewidth]{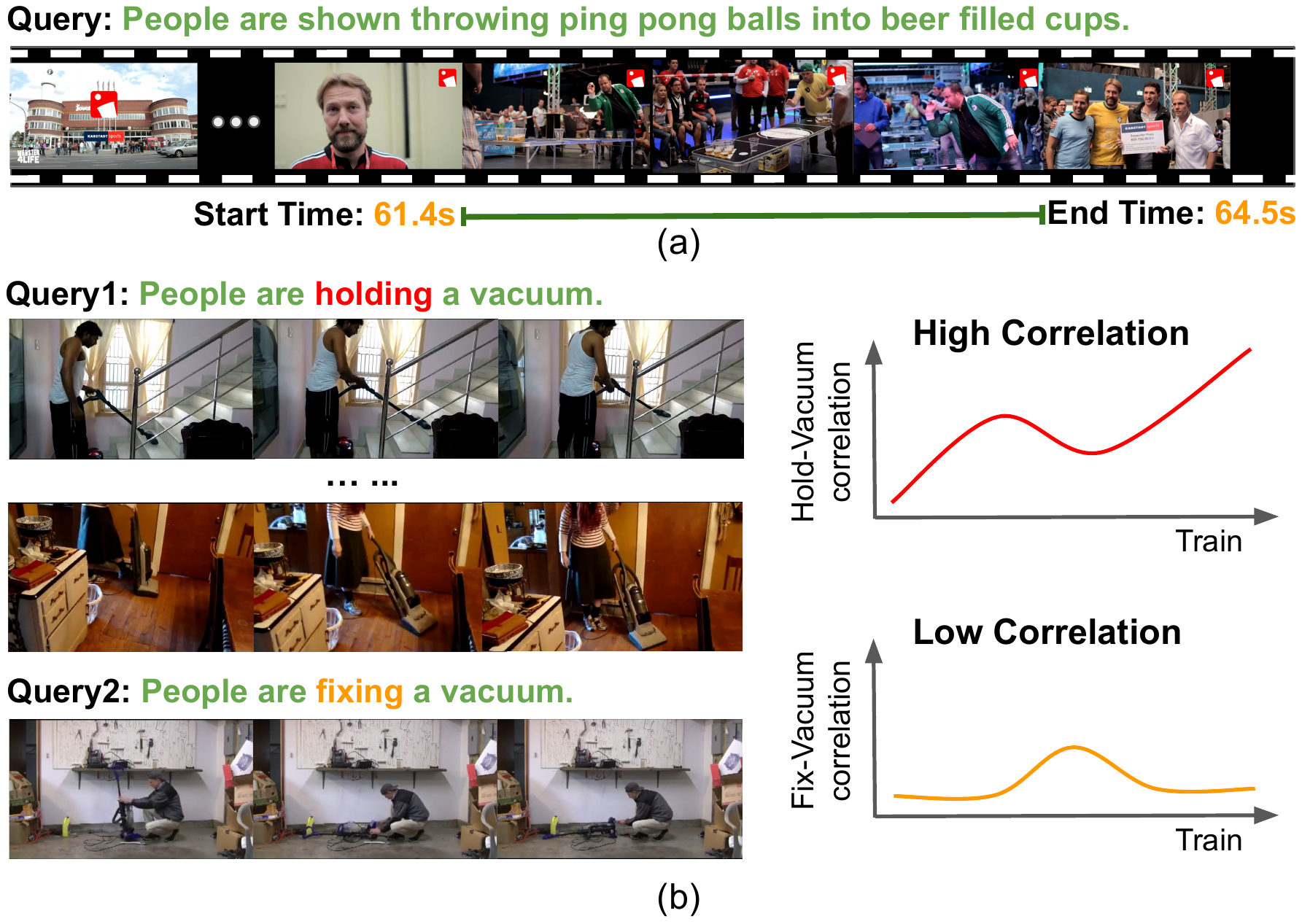}
\vspace{-4mm}
\end{center}
   \caption{(a) Illustration of video grounding. (b) Spurious correlations between object ``people'' and  ``vacuum'' and the activity ``people are holding a vacuum'' in the  Charades-TA~\cite{gao2017tall} dataset.}
\label{fig:intro}
\vspace{-4mm}
\end{figure}

Video grounding \cite{anne2017localizing,gao2017tall}, which aims to automatically locate the temporal boundaries of the target video span for a given textual description, is a challenging multimedia information retrieval task due to the flexibility and complexity of text descriptions and video content. It has been widely used in many applications, such as video question answering~\cite{lei2018tvqa} and video summarization~\cite{wu2020dynamic}. As shown in Figure \ref{fig:intro} (a), the query ``People are shown throwing ping pong balls into beer-filled cups'' involves two actions ``shown'' and ``throwing'', one role ``people'', and three objects ``ping pong balls'', ``beer'' and ``cups'', which will be located in the video with a start time (61.4s) and an end time (64.5s). To retrieve the most relevant segment, a model needs to well understand the complex interactions among these actions and objects from both language and video context, and then properly align the two sides semantic information for a prediction. 

Extensive works have been proposed for the aforementioned challenging video grounding task, which can be mainly categorized into three groups: 1) ranking methods that typically count on a two-stage propose-and-rank pipeline \cite{gao2017tall,anne2017localizing,zhang2019man}, or attention-based localization approach \cite{yuan2019find} to find out the target video span among candidates with the best matching score. 2) regression methods \cite{lu2019debug,zeng2020dense,mun2020local} that directly predict the start and end time of the target moment to avoid the heavy computations for a large number of candidates. 3) reinforcement learning \cite{wang2019language,he2019read,cao2020strong} approaches that dynamically filter out a
sequence of video frames conditioned on the given textual query and finally outputs temporal boundaries. The above studies keep pushing the boundary of state-of-the-art performance for the moment localization, and have made it possible to benefit the downstream applications \cite{lei2018tvqa,gabeur2020multi,wu2020dynamic}.

Despite the enormous success of the current neural models, we argue that these approaches may suffer from the spurious correlations between textual and visual features due to the selection bias of the dataset. As shown in Figure \ref{fig:intro} (b), a dataset involves many relevant training instances for some queries that contain the action description word ``hold'' and two objects ``people'' and ``vacuum'. Meanwhile, there are a very limited number of queries with the action ``fix'' and two objects ``people'' and ``vacuum'. We can draw a conclusion that the tokens ``people'' and ``vacuum' are highly correlated to the video moments relevant to ``people are holding a vacuum'', and have low correlations to the video moments ``people are fixing a vacuum''. Hence, a query that contains the description ``people are fixing a vacuum'' may be incorrectly located to the video segment ``people are holding a vacuum'' with high probability. We observe that such biased distributions commonly exist in many datasets, such as Charades-STA \cite{gao2017tall} and TACoS \cite{tacos2012}.

Many methods \cite{mahajan2018exploring,cui2019class,wang2020devil,zhangetal2019long} attempt to address the above issue. However, these re-sampling \cite{cui2019class} and re-weighting \cite{mahajan2018exploring} methods are mainly designed for the classification tasks, and it can be difficult for them to be applied to temporal localization. One may also consider that pre-trained knowledge is immune to the issue as they are learned from large-scale datasets. However, as highlighted by Zhang \etal \cite{yue2020interventional}, the pre-trained models, such as BERT \cite{devlin2018bert} and ResNET \cite{he2016deep}, may still suffer from the biased issue. From a very different direction, causal inference \cite{pearl2016causal} based on structured causal model (SCM) \cite{pearl2016causal} and potential outcomes \cite{rubin2019essential} have recently shown great promise, achieving state-of-the-art performance on many applications such as scene graph generation \cite{tang2020unbiased}, data clustering \cite{wang2020decorrelated}, image classification \cite{yue2020interventional}, and visual question answering \cite{agarwal2020towards,qi2020two}. Despite these success, directly applying these causal methods to the video grounding task may not yield good results, due to the more complex interactions for moment retrieval compared with the image-based bi-modal alignment. 

To address the above problem, this paper presents a novel paradigm named interventional video grounding (\texttt{IVG}) based on Pear's SCM \cite{pearl2016causal} from the perspective of causal inference. Theoretically, SCM uses the graphical formalism to treat nodes as random variables and directed edges as the direct causal dependencies between variables. We borrow the idea of the backdoor adjustment and \textit{do}-calculus theory $P(Y|do(X))$ \cite{pearl2016causal} to deconfound the aforementioned spurious correlations. The main challenge here is to get the unobserved confounders $Z$ that influences both bi-modal representations and the predictions, leading to the unexpected correlations between language and video features by only learning from the likelihood $\textit{P}(Y|X)$. Previous studies on image-based tasks treat the latent confounder as the prior knowledge or dataset distribution, and approximate them with static probabilities \cite{qi2020two,wang2020visual} of the image objects from the dataset, or the probabilities predicted by pre-trained classifiers \cite{yue2020interventional}. 
We propose a simple yet effective method to approximate the prior $P(Z)$ and then obtain $P(Y|do(X))$. Meanwhile, we introduce a dual contrastive learning method (\texttt{DCL}) to learn more informative visual representations by maximizing the MI between query and video clips to better align the bi-modal features, and the MI between the start/end time of the target moment and other clips in the video. With the above two key components \texttt{IVG} and \texttt{DCL}, our proposed \texttt{IVG-DCL} can alleviate the confounder bias and learn high-quality representations for the challenging video grounding task. Experiments on three public datasets show the effectiveness of our proposed \texttt{IVG-DCL}. Specifically, our main contributions are: 

\begin{itemize}
    \item We propose \texttt{IVG}, a novel model for video grounding by introducing causal interventions $P(Y|do(X))$ to mitigate the spurious correlations from the dataset. We also present a novel approximation method for the latent confounder based on SCM. 
    \item We propose a dual contrastive learning method \texttt{DCL} based on MI maximization to learn more informative feature representations in an unsupervised manner.
    \item We conduct quantitative and qualitative analyses on three benchmark datasets and show interesting findings based on our observations. 
\end{itemize}

\section{Related Work}
\noindent
\subsection{Video Grounding:} The task of video grounding or moment localization \cite{anne2017localizing,gao2017tall} aims to retrieve video segments for the given textual queries. 
Previous works can be mainly categorized into three groups. 1) \textbf{The ranking methods} \cite{gao2017tall,anne2017localizing,hendricks2018localizing,wu2018multi,liu2018attentive,liu2018cross,xu2019multilevel,ge2019mac,zhang2019man} rely on bi-modal matching mechanisms to obtain the target moment that has the best score. Typical works \cite{gao2017tall,anne2017localizing,xu2019multilevel,zhang2019man} in this direction resort to a two-stage propose-and-rank pipeline. These approaches highly rely on the quality of proposals and may suffer from heavy computation cost. 
2) \textbf{Regression approaches} \cite{lu2019debug,zeng2020dense,mun2020local} that regress the visual boundaries without matching scores. There are also some works \cite{chen2019localizing,ghosh2019excl,zhang2020span} that frame the video grounding as a question answering problem \cite{antol2015vqa} and treat the target moments as the answering span. 3) \textbf{Reinforcement learning methods} \cite{wang2019language,he2019read,cao2020strong} that progressively localize the targeting moment by treating the problem of temporal grounding as a sequential decision-making process. 
There also exist some other studies \cite{fan20dialog,yang2020tree,fan20persontube,gabeur2020multi,yang2020weakly,lin2020weakly,mithun2019weakly,Zhang2020CounterfactualCL} for the relevant vision-language tasks. Unlike these previous approaches, we advance the video grounding in the perspective of causal inference by mitigating the spurious correlation between video and language features. 

\subsection{Causal Inference:}
Compared to the conventional debiasing techniques \cite{zhangetal2019long,wang2020devil}, causal inference \cite{pearl2016causal,rubin2019essential,guo2020survey} shows its potential in alleviating the spurious correlations~\cite{bareinboim2012controlling}, disentangling the desired model effects~\cite{besserve2019counterfactuals}, and modularizing reusable features for better generalization ~\cite{parascandolo2018learning}. It has been widely adopted in many areas, including image classification~\cite{chalupka2014visual,lopez2017discovering}, imitation learning~\cite{de2019causal}, visual question answering (QA) ~\cite{chen2020counterfactual,abbasnejad2020counterfactual,niu2020counterfactual}, zero-shot visual recognition \cite{yue2021counterfactual}, long-tailed image recognition and segmentation~\cite{tang2020unbiased,tang2020long,zhang2020causal}, stable prediction~\cite{kuang2018stable}, policy evaluation~\cite{zou2019focused}, and treatment effect estimation~\cite{kuang2019treatment}. 
Specifically, Tang \etal \cite{tang2020long} and Zhang \etal \cite{zhang2020causal} rely on backdoor adjustment \cite{pearl2016causal} to compute the direct causal effect and mitigate the bias introduced by the confounders. Most similar to our work are DIC ~\cite{yang2020deconfounded} and CVL~\cite{abbasnejad2020counterfactual} which are proposed for image captions and image-based visual QA, and both of them adopt SCM to eliminate the bias for vision-language tasks. The key difference between our work and the previous ones is: our SCM is proposed for video-based moment localization and the latent confounder approximation considers roles, actions and objects in a moment, while DIC and CVL are designed for static image-based vision-language tasks. 

\subsection{Contrastive Learning and Mutual Information}
Contrastive learning \cite{hadsell2006dimensionality,wu2018unsupervised,zhuang2019local,misra2020self,he2020momentum,chen2020simple} methods are proposed to learn representations by contrasting positive pairs against negative pairs. Some prior works consider to maximize the mutual information (MI) between latent representations \cite{hjelm2018learning,bachman2019learning}. MI \cite{bell1995information, hyvarinen2000independent} quantifies the ``amount of information'' achieved for one random variable through observing the other random variable. There are many estimators \cite{belghazi2018mine,oord2018representation} to calculate the lower bounds for MI, which have been proven to be effective for unsupervised representation learning \cite{hjelm2018learning, velickovic2019deep,sun2019infograph}. Our \texttt{DCL} is mainly inspired by the \texttt{Deep InfoMax} \cite{hjelm2018learning}, which maximizes the MI between input data and learned high-level representations. The main differences between our work and the prior studies: 1) our \texttt{DCL} discriminates between positive samples and negative samples based on the temporal moment. 2) our dual contrastive learning module achieves two goals, i.e.,  guiding the encoder to learn better video representation, as well as better alignment between text and video.  
\section{Model}
\begin{figure}[t]
\begin{center}
\includegraphics[width=0.9\linewidth]{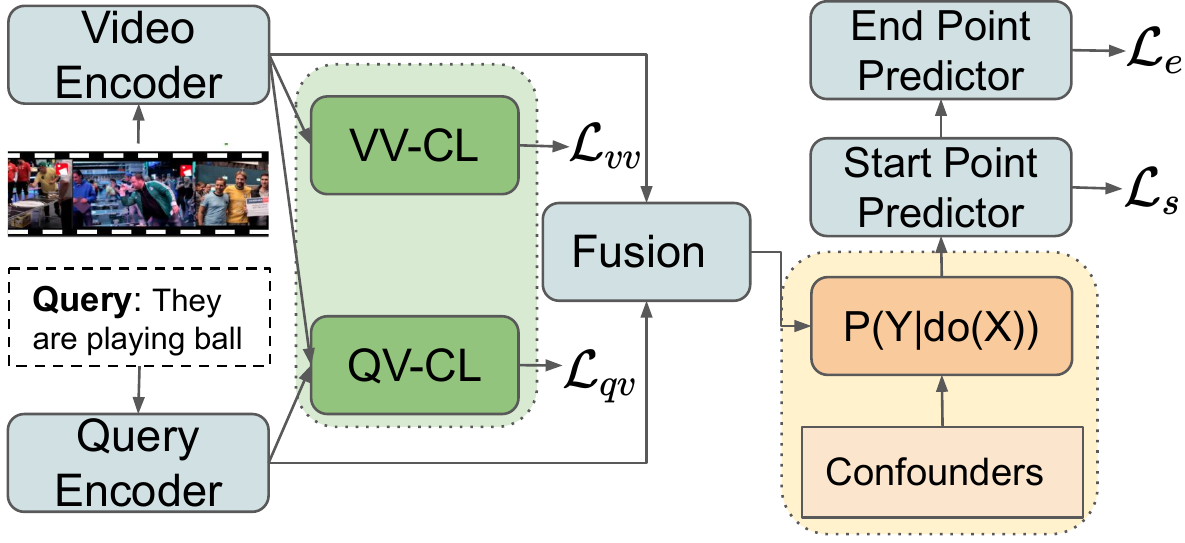}
\end{center}
\vspace{-4mm}
   \caption{The architecture of our  \texttt{IVG-DCL}. \texttt{VV-CL} and \texttt{QV-CL} refer to two contrastive modules with losses expressed as $\mathcal{L}_{vv}$ and $\mathcal{L}_{qv}$. $\mathcal{L}_s$ and $\mathcal{L}_e$ denote the cross-entropy losses for predicting the boundary of the target span.}
\label{fig:acr}
\vspace{-4mm}
\end{figure}
\subsection{Overview}
Figure \ref{fig:acr} depicts the architecture of our \texttt{IVG-DCL}. 1) Given an input query and video, the two encoders output contextualized visual and textual representations respectively. 2) Then, these representations will be fed into two contrastive modules \texttt{VV-CL} and \texttt{QV-CL} respectively to learn high-quality representations with two contrastive losses $\mathcal{L}_{vv}$ and $\mathcal{L}_{qv}$. 3) The output of two feature encoders are fed to a fusion module with a context-query attention mechanism to capture the cross-modal interactions between visual and textual features. 4) Next, to mitigate the spurious correlations between textual and visual features, we use causal interventions $P(Y|do(X))$ with event as surrogate confounders to learn representations. 
5) Finally, two losses $\mathcal{L}_s$ and $\mathcal{L}_e$ for the start and end boundaries are introduced. We use the multi-task learning paradigm to train the model with the above four different losses.

\subsection{Feature Encoder:}

\noindent
\textbf{Feature Extraction:} Let $V$ = $\{f_t\}^{\hat{T}}_{t=1}$ and $Q$ = $\{w_i\}^N_{i=1}$ be a given untrimmed video and textual query, respectively, where $f_t$ and $w_i$ denote the $t$-th frame and $i$-th word of the inputs, and $\hat{T}$ and $N$ denote the total number of frames and tokens. We use feature extractor, eg., C3D \cite{tran2015learning}, to obtain the video features $\textbf{V} = \{\textbf{v}_i\}^T_{i=1}$ $\in$ $\mathbb{R}^{T \times d_v}$, where $T$ indicates the number of extracted features  from the video, $\mathbf{v}_i$ represents the $i$-th video feature, and $d_v$ refers to the dimension of the video features. As we aim to learn to predict the boundaries of the target moment, which can be denoted as $\mathcal{T}^s$ and $\mathcal{T}^e$, we convert the boundaries $\mathcal{T}^s$ and $\mathcal{T}^e$ to the index of the video features to facilitate our learning process. Let $\mathcal{T}$ denote the video duration, the start and end time index of the moment can be expressed as $I_s = T \times \mathcal{T}^s / \mathcal{T}$ and $I_e = T \times \mathcal{T}^e / \mathcal{T}$, respectively. For the query $Q$, we also use pre-trained word embeddings, eg., Glove~\cite{Pennington2014GloveGV}, to obtain high-dimensional word features $\textbf{Q}$ = $\{\textbf{w}_i\}^N_{i=1}$ $\in$ $\mathbb{R}^{N \times d_w}$, where $\textbf{w}_i$ is the $i$-th word feature of $d_w$ dimensions. 

\noindent
\textbf{Encoders:} We use two linear layers to project the video features $\textbf{V}$ and query features $\textbf{Q}$ to the same dimension $d$. Then we refer to the previous works \cite{yuan2019find,zhang2020span} and use four convolutions layers, a multi-head attention layer, and a feed-forward layer to generate contextualized representations $\textbf{Q}’ = \{\textbf{w}'_i\}^T_{i=1}$ $\in$ $\mathbb{R}^{N \times d_v}$ and $\textbf{V}' = \{\textbf{v}'_i\}^T_{i=1}$ $\in$ $\mathbb{R}^{T \times d_v}$. The visual and textual encoders share the same parameters. 

\subsection{Contrastive Learning Module}
Contrastive loss \cite{hjelm2018learning} is able to measure the similarities between sample pairs in the representation space, and it can be served as an unsupervised objective
function for training the encoder networks with discriminative positive and negative samples. To guide the encoder to better align the textual and video representations  $\textbf{Q}’$ and $\textbf{V}'$, we treat the video clips that reside in the boundaries of the target moment as the positive samples, and the ones that are outside of the boundaries as the negative samples. And then we use a discriminative approach based on mutual information (MI) maximization \cite{velickovic2019deep} to compute the contrastive loss. In information theory, the mutual information refers to the measurement of mutual dependence between two random variables and quantifies the \textit{amount of information} obtained about one variable by observing the other variable. 

\begin{figure}[t]
\begin{center}
\includegraphics[width=0.9\linewidth]{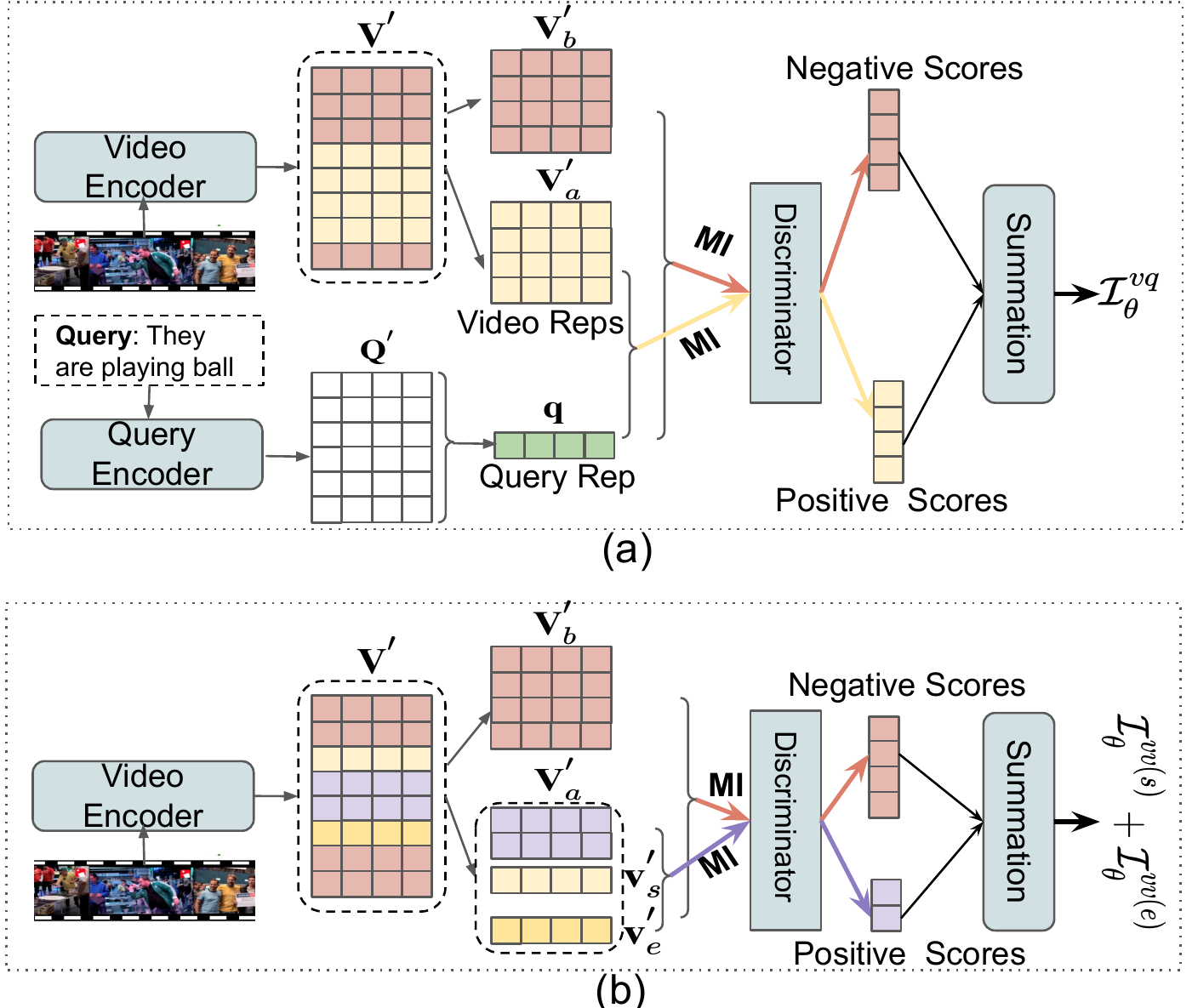}
\end{center}
\vspace{-4mm}
   \caption{Our \texttt{DCL}. $\textbf{V}'$ and $\textbf{Q}'$ refer to contextualized video and query representations, respectively. $\textbf{V}'_a$ and $\textbf{V}'_b$ represent the positive and negative samples in $\textbf{V}'$, respectively. (a) our \texttt{VQ-CL} module. We generate $\textbf{q}$ by max-pooling on $\textbf{Q}'$. (b) our \texttt{VV-CL} module. $\textbf{v}'_s$ and $\textbf{v}'_e$ refer to the video representations of start and end index of a target moment. The representations (color purple) indicate the other features in the target moment except for starting and ending point.}
\label{fig:cl}
\vspace{-3mm}
\end{figure}

\subsubsection{VQ Contrastive Learning}

Figure \ref{fig:cl} (a) presents our \texttt{VQ-CL} module. We denote $\textbf{q}$ as the query representation pooled from $\textbf{Q}'$, and $\mathcal{I}^{vq}_{\theta}$ as the MI between $\textbf{q}$ and $\textbf{V}'$ with parameter $\theta$. However, MI estimation is generally intractable for continuous and random variables. We therefore alternatively maximize the value over lower bound estimators of MI by Jensen-Shannon MI estimator \cite{hjelm2018learning}. We divide the video representation $\textbf{V}'$ into two parts, i.e., $\textbf{V}_a'$ as positive samples which denote the features that reside within a target moment, and $\textbf{V}'_b$ as negative samples that denote the features are located outside of the target segment. The MI $\mathcal{I}^{vq}_{\theta}$ can be estimated by:

\begin{equation}
\begin{aligned}
    \mathcal{I}^{vq}_{\theta}(\mathbf{q}, \textbf{V}'): = E_{\textbf{V}_a'}[sp (\mathcal{C}(\textbf{q},\textbf{V}')]  \\ -  E_{\textbf{V}_b'}[sp (\mathcal{C}(\textbf{q},\textbf{V}'))]
\end{aligned}
\label{eq:mi}
\end{equation}
where $\mathcal{C} : \mathbb{R}^{d_v} \times \mathbb{R}^{d_v} \rightarrow \mathbb{R}$ refers to the MI discriminator, and $sp(z) = log(1 + e^z)$ is the softplus function. Then we can get the contrastive loss $\mathcal{L}_{vq}$ as follows.

\begin{equation}
    \mathcal{L}_{vq} = - \mathcal{I}^{vq}_{\theta}
\end{equation}

\subsubsection{VV Contrastive Learning}

Figure \ref{fig:cl}(b) depicts the details of our \texttt{VV-CL} module. We denote $\mathcal{I}^{vv(s)}_{\theta}(\textbf{v}_s', \textbf{V}'))$ and $\mathcal{I}^{vv(e)}_{\theta}(\textbf{v}_e', \textbf{V}'))$ as the MIs between the start and end boundaries of the video and the other clips, respectively, where $\textbf{v}_s'$ and $\textbf{v}_e'$ refer to the video representations of the start and end boundaries of the target moment. We follow the same algorithm in Eq. \ref{eq:mi} to achieve the results. The loss $\mathcal{L}_{vv}$ can be expressed as follows.
\begin{equation}
    \mathcal{L}_{vv} = - \mathcal{I}^{vv(s)}_{\theta} - \mathcal{I}^{vv(e)}_{\theta}
\end{equation}
Therefore, we are able to train a better encoder with the proposed \texttt{DCL} by maximizing the MIs as follows. 
\begin{equation}
\label{eq:argmax}
    \hat{\theta} =\mathop{\arg\max}_{\theta} \mathcal{I}^{vq}_{\theta} +  \mathcal{I}^{vv(s)}_{\theta} +  \mathcal{I}^{vv(e)}_{\theta}
\end{equation}

\subsection{Fusion Module}
We leverage context-query attention (CQA) \cite{seo2016bidirectional} and follow VSLNet \cite{zhang2020span} to capture the bi-modal interaction between video and text features. We denote CQA as the function for the interaction, which takes the $\textbf{V}'$, $\textbf{Q}'$ as the inputs. The output of the fusion module $\textbf{X}$ can be expressed as:
\begin{equation}
    \textbf{X} = FFN (CQA(\textbf{V}', \textbf{Q}'))
\end{equation}
where $FFN$ refers to a single feed-forward layer. 

\subsection{Interventional Causal Model}
\begin{figure}[t]
\begin{center}
\includegraphics[width=0.87\linewidth]{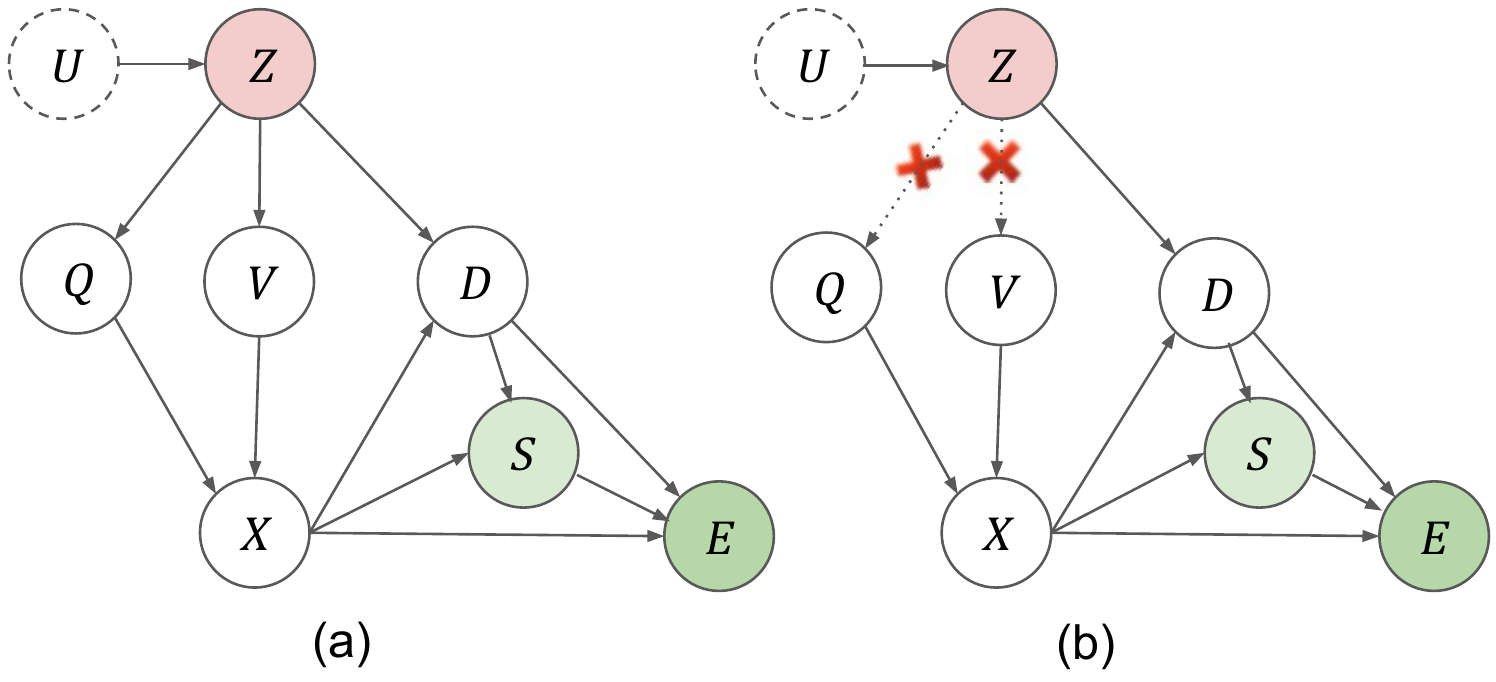}
\end{center}
\vspace{-4mm}
   \caption{The proposed interventional video grounding (\texttt{IVG}) based on structured causal model (SCM). $U$: selection bias for generating a video grounding dataset. $Z$ : the unobserved confounders that may lead to spurious correlations. $Q$: textual query. $V$: untrimmed video. $X$: the multi-modal representations. $D$: the mediator. $S$ and $E$: the start and the end boundaries of the target moment. }
\label{fig:scm}
\vspace{-4mm}
\end{figure}


\subsubsection{Structural Causal Model}
In the case of video grounding, we employ Pearl's SCM \cite{pearl2016causal} to obtain the causal effect between a query and the target moment as illustrated in Figure \ref{fig:scm} (a), where the nodes are variables and edges are causal relations. We can justify the correctness of the graph by detailed interpretation for the subgraphs as follows:

\noindent
$\boldsymbol{U \rightarrow Z \rightarrow \{V, Q\}}$. An unknown confounder $Z$ (probably imbalanced distribution of dataset caused by sampling bias $U$) may lead to spurious correlations between video clips and certain words. The $do$-operation on $\{V,Q\}$ is able to enforce their values and cuts off the direct dependency between $\{V,Q\}$ and their parent $Z$ (Figure \ref{fig:scm} (b)).

\noindent
$\boldsymbol{Z \rightarrow D \rightarrow \{S,E\}}$. Since $Z$ is the set of confounders for the dataset, we must also have $Z$ connected to prediction $\{S,E\}$ via directed paths excluding $\{V,Q\}$. This  ensures the consideration of confounding impact from $Z$ to $\{S,E\}$.  

\noindent
$\boldsymbol{X \rightarrow D \leftarrow Z}$. $D$ is a mediator that contains the specified representation of $Z$ considering the presence of $X$. The adoption of $D$ is a more flexible notation compared to directly have the edge $Z \rightarrow S$ as it permits additional effect of $X$ towards the computation of $\{S,E\}$ through $D$. This allows us to compare a variety of model architectures under the same causal framework. Besides, unlike previous models \cite{zhang2020causal} that emphasize the mediation effect of $X$ through $D$, we believe that the main constituent of $D$ should still come from $Z$, despite having only an approximation.
\subsubsection{Causal Intervention}
Conventional video grounding models, which are based on correlations between query and video, directly learn $P(S,E|Q,V)$ without considering the confounder set $Z$. In our SCM, the non-interventional prediction can be expressed using Bayes rule as:
\begin{equation}
    P (S,E|{Q,V}) = \sum_z P({S,E} | {Q,V}, z) P(z |{Q,V} ) 
\end{equation}
However, the above objective learns not only the main direct correlation from $\{Q,V\} \rightarrow X \rightarrow \{S,E\}$ but also the spurious one from the unblocked backdoor path $\{Q,V\} \leftarrow Z \rightarrow D \rightarrow \{S,E\}$. 
An intervention on $\{Q,V\}$, denoted as $do(Q,V)$, forces their value and removes the dependency from their parent $Z$. 
Therefore, we can adopt $do$-calculus and backdoor adjustment \cite{pearl2016causal} to alleviate the backdoor correlation and compute the interventional objective:
\begin{equation}
\begin{aligned}
        P (S,E|&do({Q,V})) \\
        &= \sum_z P({S,E} | do({Q,V}), z) p(z |do({Q,V}) ) \\
       &= \sum_z P({S,E} | do({Q,V}), z) p(z) \\
       &= \sum_z P({S,E} | do({X}), z) p(z) 
\end{aligned}
\end{equation}
\subsubsection{Latent Confounder Approximation}
Unfortunately, the confounder set $Z$ caused by selection bias cannot be observed directly in our case, due to the unavailability of the sampling process. To estimate its distribution, existing approaches \cite{yue2020interventional} mainly rely on image representations from pre-trained classifier, which is expensive and ignores actions embedded in video sequence. Here we propose a novel and efficient approach to approximate the confounder set distribution from the perspective of natural language. 
Our surrogate confounder set includes the vocabulary of roles, actions and objects extracted from captions, since most video clips are described as ``somebody does something''. With this in mind, we use the state-of-the-art Open Information Extraction model RnnOIE \cite{Stanovsky2018SupervisedOI} to extract verb-centered relation tuples (subject, verb, object) from caption texts and construct three sets of vocabulary accordingly. We also compute the prior probability for phrases $z$ in each set based on the dataset statistics:
\begin{equation}
	p(z) = \frac{\# z}{\sum_{i \in C}\#i}, \forall z \in C
\end{equation}
where $\#i$ is the number of occurrences of the phrase $i$ and $C$ is one of the three vocabulary sets. 
Equipped with approximated $Z$ and its prior distribution, we also propose to model the representation of $Z$ by linear projection $g$ and a word embedding layer $embed$:
\begin{equation}
	\textbf{z} = g(embed(z))
\end{equation}
while making sure $dim(\textbf{z}) = dim(\textbf{x})$,
where $\textbf{x} \in col(\textbf{X})$ and $col$ denotes the column space of a matrix and $dim$ indicates the dimension of the vector. Since $P(S,E|do(X))$ is calculated using softmax, we can apply Normalized Weighted Geometric Mean (NWGM)\cite{Xu2015ShowAA} and the deconfounded prediction can be approximated by:
\begin{equation}
\begin{aligned}
        P (S,E|do({X=\textbf{X}})) 
        &= \sum_z P({S,E} | \textbf{X} \oplus \textbf{z}) p(z) \\
        	&\approx P({S,E} | \sum_z (\textbf{X} \oplus \textbf{z}) p(z))
\end{aligned}
\end{equation}
where $\oplus$ is the broadcast add operator that adds a vector to each vector on the columns of the matrix. Finally, we can achieve the start and end index $I_s$ and $I_e$ of the target moment by the above deconfounded prediction, and also simply use cross entropy to calculate the loss for $S$ (resp. $E$) as  $\mathcal{L}_{s}$ (resp. $\mathcal{L}_{e}$). 

\subsection{Training Objectives}
The overall training loss $\mathcal{L}$ can be computed by:
\begin{equation}
    \mathcal{L} = \alpha \mathcal{L}_{vq} + \beta \mathcal{L}_{vv} + \mathcal{L}_{s} + \mathcal{L}_{e}
\end{equation}
where $\alpha$ and $\beta$ are weights for the dual contrastive losses. 
During the inference stage, the \texttt{DCL} will be ignored as it only facilitates the representation learning and requires the moment annotation to identify the contrastive samples.

\section{Experiments}

\subsection{Dataset}

\noindent
\textbf{TACoS} is constructed from MPII Cooking Composite Activities dataset~\cite{tacos2012}. We follow the same split of \cite{gao2017tall} for fair comparisons, where 10,146, 4,589 and 4,083 instances are used for training, validation, and test, respectively. Each video has 148 queries on average.

\noindent
\textbf{Charades-STA} is a benchmark for the video grounding, which is generated by~\cite{gao2017tall} based on Charades dataset~\cite{sigurdsson2016hollywood} mainly for indoor activities, with 12,408 and 3,720 moment annotations for the training and test, respectively. 

\noindent
\textbf{ActivityNet Caption} involves about 20K videos with 100K queries from the ActivityNet~\cite{caba2015activitynet}. We refer to the split used in ~\cite{zeng2020dense}, with 37,421 moments for annotations for training, and 17,505 moments for testing. For fair comparisons, we follow \cite{yuan2019find,zhang2020span} for training,  evaluation and testing.

\subsection{Experimental Settings}

\noindent
\textbf{Metrics:} We follow \cite{gao2017tall,zhang2020span,zeng2020dense} to use "R@n, IoU = $\mu $" as the evaluation metrics, which denotes the percentage of testing samples that have at least one correct result. A correct result indicates that intersection over IoU with ground truth is larger than $\mu$ in top-$n$ retrieved moments. 

\noindent
\textbf{Settings:}
We re-implement the VSLBase \cite{zhang2020span} in Pytorch and use it as our backbone network. We follow the previous works \cite{zhang2020span,zhang2020learning} to use the same pre-trained video features \cite{tran2015learning} and 300-dimension word embedding from Glove \cite{Pennington2014GloveGV}. The loss weights $\alpha$ and $\beta$ are configured as 0.1 and 0.01 respectively. 

\subsection{Model Zoo}

\noindent
\textbf{Ranking methods} rely on multi-modal matching architectures to obtain the target moment with the highest confidence score, including 2D-TAN~ \cite{zhang2020learning}, MAN~\cite{zhang2019man}, etc. Among them, 2D-TAN \cite{zhang2020learning} relies on a temporal adjacent network to localize the target moment. MAN~\cite{zhang2019man} captures moment-wise temporal relations as a structured graph and devise an adjustment network to find out the best candidate. 

\noindent
\textbf{Regression models} directly regress the moment boundaries to avoid heavy computations, including ABLR~\cite{yuan2019find}, DEBUG~\cite{lu2019debug}, DRN \cite{zeng2020dense}, and VSLNet\cite{zhang2020span}, etc. DRN \cite{zeng2020dense} relies on a dense regression network to improve video grounding accuracy by regressing the frame distances to the starting (ending) frame. VSLNet~\cite{zhang2020span} obtains the start and end indexes of the target span based on a QA framework. 


\noindent
\textbf{Reinforcement learning (RL) methods}  progressively localize the moment boundaries for a given query. 
SM-RL~\cite{wang2019language} and RWM~\cite{he2019read} treat the problem of temporal grounding as a sequential decision-making process, which naturally can be resolved by the RL paradigm \cite{ yun2017action}.

\subsection{Performance Comparisons}
The results of our method on three benchmark datasets are given in the
Table \ref{tab:charades}, Table \ref{tab:tacos}, and Table \ref{tab:activity}, respectively. As shown in Table \ref{tab:charades}, our proposed \texttt{IVG-DCL} consistently outperforms the baselines under various settings and achieves the new state-of-the-art performance on the Charades-STA dataset. Compared with VSLNet, our model achieves 3.33 points improvement in IoU=0.3 and 2.87 points measured by mIoU. Our model significantly outperforms a regression baseline VSLBase by 9.27 over IoU = 0.5 and 6.09 in terms of mIoU. The results indicate the superiority of our proposed \texttt{IVG-DCL} in predicting more accurate temporal moment.

\begin{table}[]
\begin{center}
\scalebox{0.7}{
\begin{tabular}{llccccc}
\toprule
\\
Type        & Model                                     & IoU=0.3 & IoU=0.5 & IoU=0.7 & mIoU  \\ \hline
RL          & SM-RL~\cite{wang2019language}             & -       & 24.36   & 11.17   & -     \\
            & RWM~\cite{he2019read}                     & -       & 36.70   & -       & -     \\ \hline
Ranking     & CTRL~\cite{gao2017tall}                   & -       & 23.63   & 8.89    & -     \\
            & ACRN~\cite{liu2018attentive}              & -       & 20.26   & 7.64    & -     \\
            & SAP~\cite{chen2019semantic}               & -       & 27.42   & 13.36   & -     \\
            & MAC~\cite{ge2019mac}                      & -       & 30.48   & 12.20   & -     \\
            & QSPN~\cite{xu2019multilevel}              & 54.70   & 35.60   & 15.80   & -     \\
            & 2D-TAN~\cite{zhang2020learning}           & -       & 39.70   & 23.31   & -     \\
            & MAN~\cite{zhang2019man}                   & -       & 46.53   & 22.72   & -     \\
            \hline
Regression  & DEBUG~\cite{lu2019debug}                  & 54.95   & 37.39   & 17.69   & 36.34 \\        
            & DRN~\cite{zeng2020dense}                  & -       & 45.40   & 26.40   & -     \\
            & VSLBase~\cite{zhang2020span}              & 61.72   & 40.97   & 24.14   & 42.11 \\
            & VSLNet~\cite{zhang2020span}               & 64.30   & 47.31   & 30.19   & 45.15 \\
           \hline \hline
           & \textbf{Ours}                              & \textbf{67.63}       & \textbf{50.24}       & \textbf{32.88}       & \textbf{48.02}     \\ \bottomrule
\end{tabular}
}
\end{center}
\vspace{-3mm}
\caption{Performance comparisons on the Charades-STA dataset. Note that we do not finetune the feature extractor.}
\label{tab:charades}
\vspace{-3mm}
\end{table}
\begin{table}[]
\begin{center}
\scalebox{0.7}{
\begin{tabular}{llccccc}
\toprule
\\
Type        & Model                              & IoU=0.1 & IoU=0.3 & IoU=0.5 & IoU=0.7 & mIoU  \\ \hline
RL          & SM-RL~\cite{wang2019language}      & 26.51   & 20.25   & 15.95   & -       & -     \\ 
            & TripNet~\cite{hahn2019tripping}    & -       & 23.95   & 19.17   & -       & -     \\
            \hline
Ranking     & ROLE~\cite{liu2018cross}           & 20.37   & 15.38   & 9.94    & -       & -     \\
            & MCN~\cite{anne2017localizing}      & 14.42   & -       & 5.58    & -       & -     \\
            & CTRL~\cite{gao2017tall}            & 24.32   & 18.32   & 13.30   & -       & -     \\
            & ACRN~\cite{liu2018attentive}       & 24.22   & 19.52   & 14.62   & -       & -     \\
            & QSPN~\cite{xu2019multilevel}       & 25.31   & 20.15   & 15.23   & -       & -     \\
            & MAC~\cite{ge2019mac}               & 31.64   & 24.17   & 20.01   & -       & -     \\
            & SAP~\cite{chen2019semantic}        & 31.15   & -       & 18.24   & -       & -     \\
            & TGN~\cite{chen2018temporally}      & 41.87   & 21.77   & 18.90   & -       & -     \\
            & 2D-TAN~\cite{zhang2020learning}    & 47.59   & 37.29   & 25.32   & -       & -     \\
            \hline
Regression  & SLTA~\cite{jiang2019cross}         & 23.13   & 17.07   & 11.92   & -       & -     \\
            & VAL~\cite{song2018val}             & 25.74   & 19.76   & 14.74   & -       & -     \\
            & ABLR~\cite{yuan2019find}           & 34.70   & 19.50   & 9.40    & -       & -     \\
            & DEBUG~\cite{lu2019debug}           & -       & 23.45   & 11.72   & -       & 16.03 \\
            & DRN~\cite{zeng2020dense}           & -       & -       & 23.17   & -       & -     \\
            & VSLBase~\cite{zhang2020span}       & -       & 23.59   & 20.40   & 16.65   & 20.10 \\
            & VSLNet~\cite{zhang2020span}        & -       & 29.61   & 24.27   & \textbf{20.03}   & 24.11 \\
            \hline \hline
           & \textbf{Ours}                      & \textbf{49.36}        & \textbf{38.84}   & \textbf{29.07}   & 19.05   & \textbf{28.26} \\ \bottomrule
\end{tabular}
}
\end{center}
\vspace{-4mm}
\caption{Performance comparisons on the TACoS dataset.}
\label{tab:tacos}
\vspace{-3mm}
\end{table}

\begin{table}[]
\begin{center}
\scalebox{0.7}{
\begin{tabular}{llccccc}
\toprule
\\
Type                            & Model                                     & IoU=0.3 & IoU=0.5 & IoU=0.7 & mIoU  \\ \hline
RL                              & TripNet~\cite{hahn2019tripping}           & 48.42   & 32.19   & 12.93   & -     \\
                                & RWM~\cite{he2019read}                   & -       & 36.90   & -       & -     \\
                                 \hline
Ranking                         & MCN~\cite{anne2017localizing}             & 39.35   & 21.36   & 6.43    & -     \\
                                & TGN~\cite{chen2018temporally}             & 43.81   & 27.93   & -       & -     \\
                                & CTRL~\cite{gao2017tall}                   & 47.43   & 29.01   & 10.34   & -     \\
                                & ACRN~\cite{liu2018attentive}              & 49.70   & 31.67   & 11.25   & -     \\ 
                                & QSPN~\cite{xu2019multilevel}              & 52.13   & 33.26   & 13.43   & -     \\
                                & ABLR~\cite{yuan2019find}                  & 55.67   & 36.79   & -       & 36.99 \\
                                & 2D-TAN~\cite{zhang2020learning}           & 59.45   & \textbf{44.51}   & 26.54   & -     \\
                                 \hline
Regression                      & DRN~\cite{zeng2020dense}                  & 45.45   & 24.36   & -       & -     \\
                                & DEBUG~\cite{lu2019debug}                  & 55.91   & 39.72   & -       & 39.51 \\
                                & VSLBase~\cite{zhang2020span}              & 58.18   & 39.52   & 23.21   & 40.56 \\ 
                                & VSLNet~\cite{zhang2020span}               & 63.16   & 43.22   & 26.16   & 43.19 \\\hline \hline
                                & \textbf{Ours}                             & \textbf{63.22}       &43.84        &\textbf{27.10}        &\textbf{44.21}      \\ \bottomrule
\end{tabular}
}
\end{center}
\vspace{-3mm}
\caption{Performance comparisons on ActivityNet Caption.}
\vspace{-3mm}
\label{tab:activity}
\end{table}

Table \ref{tab:tacos} summarizes the comparisons on the TACoS and it shows that our \texttt{IVG-DCL} performs best among all baselines under IoU=0.1, IoU=0.3, IoU=0.5 and mIoU. For example, our \texttt{IVG-DCL} model outperforms a strong baseline 2D-TAN and the recently proposed DRN by 3.75 and 5.9 in IoU=0.5. The results further confirm the superiority of our approach in capturing the complex interactions of roles, actions, and objects in the video grounding task. Compared to VSLNet, our \texttt{IVG-DCL} achieves comparable results on IoU=0.7, and beats it on other settings with significant improvements, e.g., 9.23 gains over IoU=0.3.

Table \ref{tab:activity} reports the results on the ActivityNet Caption. We observe that our model achieves state-of-the-art results in most settings. However, the performance gain on this dataset is much smaller than the previous two datasets. One possible reason is that the activities on the ActivityNet Caption are much more complex with more roles, actions, and objects, and the target moment duration is much longer. For example, the average duration of the target moment is 36.18s \cite{zhang2020span}, while the ones on Charades-STA and TACoS are 8.22s and 5.45s, respectively. 

\subsection{Comparisons with Causal Models}
\begin{table}[]
\begin{center}
\scalebox{0.95}{
\begin{tabular}{lcccc}
\toprule
Causal Model & IoU=0.3 & IoU=0.5 & IoU=0.7 & mIoU  \\
\midrule
FWA          &   60.67  &   40.99  &   22.85  & 41.70 \\
CWA          &   64.35  &   48.60  &   30.30  & 45.63 \\
VCR-CNN      &   65.62  &   47.58  &   28.47  & 45.94 \\
\midrule
IVG(Ours)    &   67.63  &   50.24  &   32.88  & 48.02 \\
\bottomrule
\end{tabular}
}
\end{center}
\vspace{-3mm}
\caption{Comparisons with different causal models on the Charades-STA dataset.}
\label{tab:causal_charades}
\vspace{-1mm}
\end{table}

\begin{table}[]
\begin{center}
\scalebox{0.8}{
\begin{tabular}{lccccc}
\toprule
Causal Model & IoU=0.1 & IoU=0.3 & IoU=0.5 & IoU=0.7 & mIoU  \\
\midrule
FWA          & 48.34  &   36.87  &   26.19  &   16.17  & 26.47 \\
CWA          & 49.59  &   36.37  &   27.07  &   17.62  & 27.18 \\
VCR-CNN      & 47.91  &   36.87  &   27.27  &   16.97  & 26.76 \\
\midrule
IVG(Ours)    & 50.99  &   38.79  &   28.89  &   18.25  & 28.35 \\
\bottomrule
\end{tabular}
}
\end{center}
\vspace{-3mm}
\caption{Results of different causal models on the TACoS.}
\vspace{-3mm}
\label{tab:causal_tacos}
\end{table}

\begin{figure*}
    \centering
    \includegraphics[width=385pt]{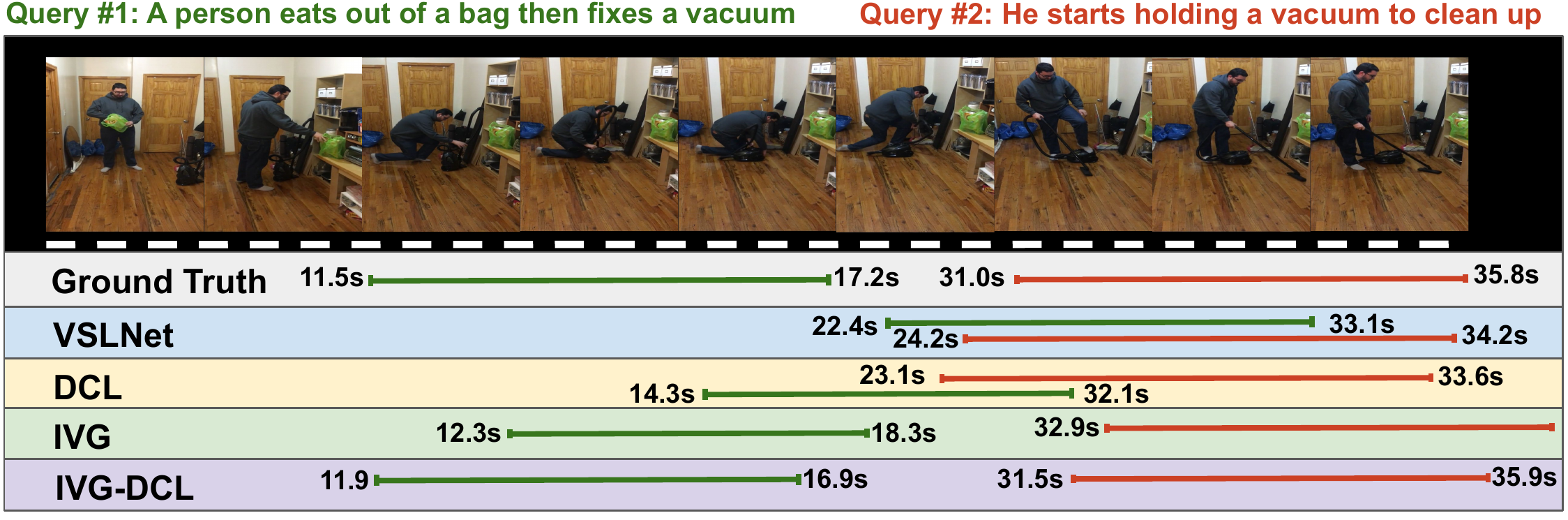}
    \vspace{-1mm}
    \caption{A case study on the Charades-STA dataset to demonstrate the capability of our model in mitigating the spurious correlations between textual and video features.}
    \label{fig:case}
    \vspace{-4mm}
\end{figure*}

We also adapt four causal models from the image-based tasks and compare them with our proposed \texttt{IVG-DCL}, including feature-wise adjustment (FWA) \cite{yue2020interventional}, class-wise adjustment (CWA) \cite{yue2020interventional} and VCR-CNN \cite{wang2020visual}. The detailed descriptions are given as follows.
1) FWA divides the final feature representation to disjoint subsets with respect to the dimensions and treats the confounder as feature subset selector. 2) Primarily designed for image classification, CWA uses image classes as data stratum to approximate the confounder distribution. Since we entirely rely on pre-trained video features in our video grounding task, we, therefore, adjust this technique for object phrases and compute average word embedding of the object phrases as ``class'' representation. 3) VCR-CNN is similar to class-wise adjustment but adds an attention-based method to combine the representation between \textit{X} and the confounding variable. As shown in Table \ref{tab:causal_charades} and Table \ref{tab:causal_tacos}, our proposed causal model outperforms these baselines on both Charades-STA and TACoS under various video grounding settings. We believe that the improvement stems from our novel latent confounder approximation method, which considers the roles, actions, as well as objects.  These factors are important clues to distinguish  activities. The comparisons confirm our hypothesis that such a novel design is useful for the video grounding task. Whereas the previous image-based approaches only consider objects in SCMs.    


\subsection{Ablation Study}
The ablation study is given in Table \ref{tab:ablation} to illustrate the contributions of each component of our model. As shown, removing the proposed \texttt{IVG} module will result in 4 points of performance drop on average, indicating that the causal model plays the key role to the performance gain. We also observed that the removal of any contrastive learning module, i.e., \texttt{QV-CL} or \texttt{VV-CL} will decrease  the performance by 1 point on average. While the removal of \texttt{DCL} will bring about 2 points of performance drop on average. This indicates that both contrastive modules contribute to the temporal localization, and combing these two will bring more gains. It is not surprising that removing both \texttt{IVG} and \texttt{DCL} will sharply decrease by 10 points on average, as essentially it becomes the VSLBase model. 
\begin{table}[h]
\begin{center}
\scalebox{0.94}{
\begin{tabular}{lcccc}
\toprule
Model & IoU=0.3 & IoU=0.5 & IoU=0.7 & mIoU  \\
\midrule
Full Model         & 67.63   & 50.24   & 32.88   & 48.02     \\ 
w/o IVG          & 64.70   & 47.60   & 30.80   & 45.34     \\ 
w/o QV-CL          & 66.30   & 49.31   & 31.19   & 47.15 \\
w/o VV-CL      & 66.75       &49.16   & 32.17   & 47.24     \\
w/o DCL      &65.21        & 48.76   & 31.93   & 46.94     \\
w/o IVG+DCL      & 61.72 & 40.97   & 24.12   & 42.11     \\
\bottomrule
\end{tabular}
}
\end{center}
\vspace{-3mm}
\caption{Ablation study on the Charades-STA dataset.}
\vspace{-3mm}
\label{tab:ablation} 
\end{table}

\subsection{Sensitivity on $\alpha$ and $\beta$}
To understand how the loss weights $\alpha$ and $\beta$ influence the performance, we manually set different values to the two hyper-parameters. Figure \ref{tab:sensitity} reports the results. We found that the best combination is $\{\alpha=0.1, \beta=0.01\}$. These weights indicate the importance of each loss in our multi-task learning paradigm for temporal localization. We found that increasing the $\alpha$ and $\beta$ will lead to performance drop, and this aligns with our hypothesis that the \texttt{IVG} plays more important role compared with \texttt{DCL}. We also observed that the combination $\{\alpha=1, \beta=1\}$ will even decrease the performance by 5 points on average. This implies that directly integrating unsupervised representation learning to SCMs may not yield good performance. 

\begin{table}[h]
\begin{center}
\scalebox{0.94}{
\begin{tabular}{lcccc}
\toprule
$\{ \alpha, \beta\}$ & IoU=0.3 & IoU=0.5 & IoU=0.7 & mIoU  \\
\midrule
$\{ 0.1, 0.01\}$         & 67.63   & 50.24   & 32.88   & 48.02     \\ 
$\{ 1, 1\}$          &   62.93  &   43.49  &   26.08  & 43.65 \\
$\{ 0.5, 0.1\}$         &   64.44  &   49.97  &   31.45  & 46.09 \\
$\{ 0.1, 0.5\}$      &   63.23  &   44.68  &   27.37  & 44.08 \\
$\{ 1.5, 1\}$         &   64.33  &   44.78  &   27.04  & 44.89 \\ 
\bottomrule
\end{tabular}
}
\end{center}
\vspace{-3mm}
\caption{Sensitivity analysis on the Charades-STA dataset.}
\label{tab:sensitity} 
\vspace{-4mm}
\end{table}

\subsection{Qualitative Analysis}
We show a case in Figure \ref{fig:case} from Charades-STA dataset to demonstrate how our proposed \texttt{ICV-DCL} alleviates the spurious correlation between text and video features. As there are a lot more relevant training instances for query $\#$2 ``he starts holding a vacuum to clean up'' compared to query $\#$1 ``a person fixes a vacuum'' (208 instances vs. 35 instances) in the training set, there will exist unexpected high correlations between vacuum-cleaning queries and vacuum-related video moments. The conventional VSLNet that is trained only based on the correlations will tend to incorrectly localize the query $\#1$ to the query $\#2$ related segments. We observe that our \texttt{IVG-DCL} outputs more accurate retrieval for both two queries as it is able to better distinguish between ``holding'' and ``fixing''. We believe such improvement mainly stems from the proposed causal model.

\section{Conclusion}
This paper proposes \texttt{IVG-DCL}, a novel model which aims to eliminate the spurious correlations between query and video features based on causal inference. Experiments on three standard benchmarks show the effectiveness of our proposed method. In the future, we plan to extend our approach to more unbiased vision-language tasks that involve dialogues \cite{lei2018sequicity,wang2019persuasion,le2019multimodal} and question answering \cite{antol2015vqa,yu2019deep,chen2020counterfactual}.  
\section*{Acknowledgments}
\vspace{-1mm}
We would like to thank the anonymous reviewers for their helpful comments.
This research is supported by the National Research Foundation, Singapore under its AI Singapore Programme (AISG Award No: AISG-RP-2019-012), and partially supported by SUTD Project (No: PIE-SGP-Al2020-02), the Agency for Science, Technology and Research (A*STAR) under its AME Programmatic Fund (Project No: A18A1b0045 and No: A18A2b0046). Any opinions, findings and conclusions or recommendations expressed in this material are those of the authors and do not reflect the views of National Research Foundation, Singapore and AI Singapore.

{\small
\bibliographystyle{ieee_fullname}
\bibliography{cvpr}
}

\end{document}